# FaceBots: Steps Towards Enhanced Long-Term Human-Robot Interaction by Utilizing and Publishing Online Social Information

Nikolaos Mavridis*, Shervin Emami*, Chandan Datta*, Wajahat Kazmi*,
Chiraz BenAbdelkader†, Panos Toulis‡, Andry Tanoto§, Tamer Rabie*
*Interactive Robots and Media Lab, CIT, Maqam Campus, UAE University
Al Ain, United Arab Emirates, Email: irmluaeu@gmail.com
†New York Institute of Technology, Abu Dhabi, United Arab Emirates
Email: chiraz@nyit.edu
‡Aristotle University, Thessaloniki, Greece
Email: ptoulis@olympus.ee.auth.gr
§Heinz-Nixdorf Institute, University of Paderborn, Paderborn, Germany
Email: Andry.Tanoto@hni.uni-paderborn.de

*Abstract*— Our project aims at supporting the creation of sustainable and meaningful longer-term human-robot relationships through the creation of embodied robots with face recognition and natural language dialogue capabilities, which exploit and publish social information available on the web (Facebook). Our main underlying experimental hypothesis is that such relationships can be significantly enhanced if the human and the robot are gradually creating a pool of shared episodic memories that they can co-refer to ("shared memories"), and if they are both embedded in a social web of other humans and robots they both know and encounter ("shared friends"). In this paper, we are presenting such a robot, which as we will see achieves two significant novelties.

## I. INTRODUCTION

The main problem addressed by this project is that of the creation of sustainable and meaningful long-term human robot relationships. This is a most important problem towards our ultimate goal of human-robot symbiosis, i.e. harmonious and mutually beneficial living together of the two species. In the shorter term, this is an important problem towards the successful application of robots to numerous areas: disabled and elderly assistance / companionship, supporting education, and more. So far, empirical investigations have shown that we have not advanced significantly yet towards its solution: Although existing robotic systems are interesting to interact with in the short term, it has been shown that after some weeks of quasi-regular encounters, humans gradually lose their interest, and meaningful longer-term human-robot relationships are not established. For example, in the case of Robovie [1], there was a steady and significant decrease in the total time of interaction of the robot with humans over six months - interest had worn off. Our proposed solution to the problem of creating sustainable and meaningful long-term human robot relationships is based on an underlying hypothesis: That such relationships can be significantly enhanced if the human and the robot are gradually creating a pool of shared episodic memories that they can co-refer to ("shared memories"), and if they are both embedded in a social web of other humans and robots they both know and encounter ("shared friends"). Thus, here we present a conversational mobile robot with face recognition that is connected to Facebook, a highly successful online networking resource for humans, towards enhancing longer-term human robot relationships, by helping to address the above two prerequisites. The contribution to the field of the project is expected to be significant. Apart from many tangential side-gains elaborated in the discussion section, our system achieves two important novelties: being the first such robot that is embedded in a social web, and being the first robot that can purposefully exploit and create social information that is available online. Furthermore, it is expected to provide empirical support for our main driving hypothesis, that the formation of shared episodic memories within a social web can lead to more meaningful long-term human-robot relationships. The experience gained by the creation of such a system as well as the software created is invaluable towards providing similar capabilities to other robots, and as a starting point for further enhancements of robots truly embedded in a social web that use and create online social information. Finally, the exposure of the robot to Facebook, through the public availability of its own Facebook page containing its friends and experiences as well as photos, will create public interest that will further support endeavours to similar directions in the future.

## II. RELATED RESEARCH

Although numerous attempts towards interactive social robots have taken place (Kismet [2], Leonardo [3], Maggie [4], Robovies [5] and more), no existing systems have utilized a connection between robots and Facebook. However, face-detecting conversational robots are not new; there are numerous projects built-around face-detecting robots [6],[7], which might even carry out conversations with multiple humans, such as in [8]. Regarding the sustainability of human-robot relationships, a key long-term (six month) study is [1]. Shorter field studies in other contexts have taken place in the past; for example the 18-day field trial of conversational robots in a Japanese elementary school [9]; and numerous are underway, including a possible massive deployment of humanoids in malls [Ishiguro, personal communication]. Finally, regarding the real-time utilization of web resources by robots, much has not been done yet, but exciting prospects exist; for example, "Peekaboom" [10], could serve as a real-time repository for object recognition.

## III. SYSTEM ARCHITECTURE

### A. Hardware

Our robot is composed of an ActivMedia PeopleBot robot [11], augmented with a SICK laser range finder, a touch screen, and a stereo Bumblebee camera [12] on a pan-tilt base [13] that is at eye-level with humans.

### B. Software Overview

We have created an expandable modular software architecture, with modules intercommunicating through the ICE IPC system [14]. The modules can be running on multiple CPUs or PCs which are part of a network, and are written in C++, Java, and Perl. Effectively, a callable-method API is exposed by each module towards the others. The modules we have created are: (M1) Vision Module with Face Detection & Recognition, from camera- or Facebook- derived pictures. Includes real-time externally callable training set modification / new classifier generation capabilities, and pluggable face detectors / classifiers. (M2) Natural Language Dialogue Module, with real-time language model switching capabilities. (M3) Social Database Module, which locally holds basic personal info / friendship relationship / simple event data / photos for the people the robot knows, and which connects and updates through Facebook for those that are members of it. (M4) Navigation and Motion Module, to build a map of its environment and drive to key social locations, and (M5) Controller Module, which issues calls to all other modules, and where high-level system operation routines can easily be scripted. A more detailed description of the modules follows.

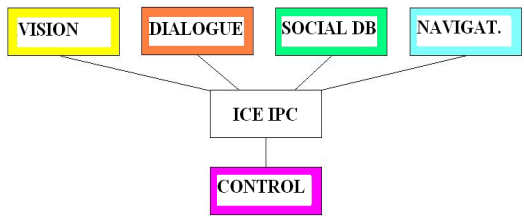

Fig. 1. Modules intercommunicating using ICE IPC

### C. Vision Module

The purpose of the vision module is to detect human faces and recognize their identity, either using images fed by be the robot's onboard camera, or using Facebook-derived photos, which are passed to it through the social database module. The module is written in C++. Its API provides externally callable methods for doing the following online tasks:

Create/add/remove face training set (for given person)
Create/remove/retrain face classifier (for given person)
Export a camera snapshot or detected face regions
Export recognized identities (score vector or hard decision)
Tagged Image handling (for Facebook-imported photos)
Set/query operational parameters

*1) Capture:* At the front end of the module, there is a camera-capture connection that first looks for the onboard stereo camera, and if this is not found, looks for a remote stereo camera coming through an ICE feed.

*2) Face Detection & Preprocessing:* Face detection follows after the capture step. This module is based on the OpenCV implementation of [15]. The detected rectangular regions which are candidate faces, are then passed through a simple skin color detection algorithm. This algorithm is based on an RGB-triplet inequality, which classifies each pixel as either skin-color or non-skin. Then, the percentage of skin color pixels in the candidate face region is calculated, and if the percentage is less than the experimentally chosen threshold of 20%, the candidate face region is discarded. Once a candidate face region passes the skin filtering test, an elliptical mask is applied to the region in order to remove irrelevant background clutter, followed by brightness normalization. The latter consists of histogram equalization and normalizing pixel values to have a zero mean and unit standard deviation [16]. At the end of these multiple processing phases, an elliptical face region is obtained.

*3) Face Recognition & Temporal Evidence Accumulation:* The elliptical face region is then fed to an embedded-HMM-based array of classifiers [17], one classifier for each person in our face database. The output of each of the embedded-HMM classifiers is a scaled log-probability measure of how well the currently viewed face matches the classifiers model. This is furthermore accumulated through a moving-window-average running across multiple frames. The choice of the number of frames (window size) and the evidence accumulation method is discussed in the tuning and evaluation section of this paper. Then, the variance of the temporally-accumulated vector of scores is calculated, and if it is below a threshold (i.e. if all the classifiers are almost equally confident about the identity of the face), then the face is marked as unknown. The choice of this threshold will also be discussed later, in the tuning and evaluation section.

*4) Facebook Photo Handling:* Facebook photos often come pretagged. However, the tags supplied contain a name augmented with a rough estimated center location, and not with a rectangular bounding box. Therefore, for a pre-tagged face, face detection is still applied, and the detected face region whose center is closest to the rough estimated center reported by Facebook is chosen as the bounding rectangle. One further complication arises because the user-tagged faces in Facebook photos might be either in frontal or in profile poses. However, we currently cannot recognize profile faces but only frontal, and thus we need to make sure that a user-tagged face in profile pose will not be ignored while a nearby frontal face is detected. In this unfortunate but quite usual case, the bounding rectangle of a nearby face of another person might be reported from our module. Even worse, it might subsequently be used as a training set picture for the classifier of the user-tagged person, while it belongs to somebody else. Therefore, we perform face detection with two face detectors in parallel: a profile-tuned detector as well as a frontal-tuned detector (while, as we mentioned before, we only have frontal face, and not profile, face recognizers). If the nearest face to the user-supplied rough center is in profile pose, then it is discarded, and so even if the second-nearest might be frontal, it is not reported. However, if the nearest face to the user-supplied rough center is in frontal pose, then it is reported, and thus can be safely used as a training set picture.

*5) Training sets:* One of the interesting novelties of our system is that it has access to two groups of pictures - coming either from Facebook photos (fully tagged, partially tagged, or untagged), or from live or stored camera pictures. The important question of appropriate training set selection, pruning, and retraining is further discussed in the tuning and evaluation section.

### D. Natural Language Dialogue Module

The purpose of the natural language dialogue-support module is to provide speech recognition, speech synthesis, as well as basic NLP services. Speech recognition is based on Sphinx 4 [18], and language models can be switched during operation. Speech synthesis on based on the Cepstral text-to-speech system that is part of the ARIA SDK [19] of the Peoplebot robot. The module is written in a mix of Java and C++.

### E. Social Database Module

The purpose of the social database module is to locally store relevant social information for the friends of the robot, and to perform acquisition and deposition of social information from Facebook. The module contains two databases: the social database and the interaction database. The social database contains entries both for people that the robot encounters which are on Facebook (Facebook friends), as well as for people that the robot encounters which are not (non-Facebook friends). Encounters can be either physical (face-to-face) or virtual (online). Furthermore, the module also contains an interaction database, storing important past interactions that can be referred to. The ultimate purpose of the social information obtained and deposited is to enable meaningful and interesting interactions between the robot and its friends, and we will soon elaborate more on how this is achieved. The database is implemented in MySQL, and the module uses interoperable Java and Perl objects, and also utilizes the Facebook developers API.

*1) Social Database Structure:* This module is essentially a friend database, which includes a subset of the Facebook-available information. It contains fields for general information (affiliation, current location, education, highschool, hometown location, work history), a friend list (containing the list of friends of the robot's friend whose entry we are describing), an event list (as posted on Facebook), and a set of photos possibly with tags.

*2) Interaction Database Structure:* This module is essentially a form of an episodic memory of the robot, whose primary key is a timestamp. Every session of interactions with a specific friend has a unique ID. Also, other fields include an interaction type identifier and a description, as well as a number of boolean flags, and the userID of the friend the robot is interacting with.

*3) The Social Module API:* methods for accessing the social database (and Facebook) include general information reads / writes, friend-list derived relation queries (including some higher-order queries which are however constrained by privacy / visibility settings), event-list queries as well as methods for getting / depositing possibly tagged photos. On the other hand, there are also methods for direct messaging through Facebook, and for accessing the interaction database. The former include methods for messaging and chatting, while the latter include methods for depositing and selectively retrieving past interactions (including for example past responses of the user to the robot's questions).

### F. Navigation and Motion Module

The purpose of this module is to provide basic navigation, mapping, as well as motion capabilities to our system. The module is written in C++, and is using the ARIA navigation library (ARNL). Its API currently provides externally callable methods for:
Start/Stop spatial exploration
Report current position / GoTo target position
Move camera pan-tilt unit

### G. Controller Module

The purpose of the central controller module is to orchestrate the operation of the facebot by issuing calls to the other modules. Within it, high-level system routines can easily be scripted, in a form that hides most of the module-specific implementation details.

## IV. SYSTEM OPERATION

The most important underlying question regarding the design of the high-level system operation (scripted within the main controller module) is: how can one appropriate utilize social information and interaction history information in order to create interesting interactions? Here, a number a possibilities was initially explored, starting with single-initiative robot-driven dialogue. Some examples from an actual demo are given below:

*Greetings / Recognition:*
H:Hello! My name is Sarah. Welcome to the interactive robots and media laboratory
R:Hi! Are you Panos Toulis?
H:Yes
(Robot changes its online status to "interacting with Panos Toulis")

*Query of overall state:*
R: Hey, Panos Toulis, are you doing well today?
H: Yes, pretty well

*General news-based discussion*
R: Let me tell you some interesting news that I've heard
R: The latest windows release will be available to everyone after a surge in demand crashed the microsoft website on the 9th of january
(item taken from BBC news online)
R: Have you heard about this?
H: Yes
R: That's great!

*Robot discusses facebook status of human*
R: I see you are chasing robots in the desert

*Robot gives news about a mutual friend, acquired through facebook*
R: Did you know that our friend Shervin Emami has posted a new photo on facebook?
(information acquired from facebook minifeed)
H: No

*Robot sends reminder to human*
R: I am sending you a message about this to check it out

*Robot remembers past encounter with mutual friend*
R: I saw Shervin yesterday evening. He was doing well.
(information acquired from interaction database)

*Robot gives news about a mutual friend, acquired through facebook*
H: Did you know that our friend Chandan is "waiting for sarah to have that little sweet conversation with me"?
(information acquired through chandan's facebook status)
R: Yes, of course.

*Robot connects an online friend with the human*
R: One of our friends, Nikolaos Mavridis, is online - would you like me to send him a message?
H: Yes
('Nikolaos, Panos says hello" is sent)

*End of interaction*
R: Hey, Panos, it was nice talking to you!
R: I have to go now.
R: See you later!

A generic framework systematizing the utilization of social information and structuring the dialogues is under development. The overall form of the actual demo a session of which was analyzed above is the following: The robot initially wanders around our lab, avoiding obstacles, and occasionally speaking to itself. When a human face is detected through the vision system, an attempt towards recognition is made, and if there is enough confidence about identity, the robot greets the human and asks if he is indeed the person the robot has recognized. If not, the second choice is announced, and a verification question is given again. Then, some pictures are taken, which are added to the training set of the appropriate classifier: either of the already known recognized person, or of a new classifier in case of a new person, who is also asked about his name.

In case of a new person, there is an attempt to find social information about him/her through Facebook, if he/she already is a member. For example, if the new person according to Facebook is a friend of an already known friend of the robot, then this is announced and indirectly asked for confirmation. In the case of an already known person appearing before the robot and being saluted and recognized correctly, a mix of the basic dialogue steps exemplified by the above transcript is utilized. For example, status changes of mutual friends are discussed, news items announced, reference is made to possible meetings with common friends in the mean time or to previous meetings with the person, instant messaging to other online friends mediated by the robot takes place etc. During the interaction, information is also posted on the robot's facebook page. Also, some pre-scripted segments of dialogue, containing announcements or jokes, can embellish the conversations. Finally, the robot says goodbye and continues its wandering. An earlier demo of the robot is already published as a video accepted by the HRI 2009 conference.

## V. TUNING AND EVALUATION

In such a complex system, there exist multiple parameters that need to be tuned as well as many discrete design choices to be taken. Furthermore, numerous types of evaluation can be carried out. Some of these are:
T1) Module-level evaluations: what is the performance of the vision system, of speech recognition etc. when viewed as isolated modules.
T2) System-level evaluations: engineering metrics implicating more than one module - for example delay between appearance of face and announcement of greeting, crossing across the vision, main controller, and speech modules.
T3) Task-centered evaluations: successful task completion rates, statistics of task duration, task-performance metrics, quantitative error analysis and error taxonomies etc.
T4) User-centered evaluations: self-reported or externally measured, including user satisfaction, ease-of-use ratings, expectations etc.
T5) Human-replacement-oriented evaluations: how close were the actions (speech and motor) taken by the robot to those of a human when put in the same situational context.

T6) Long-term field trials: for example, measurements of frequency, duration, and content of interactions during a multi-month operational deployment.

Here we will present some first results belonging mainly to the first, second and third type - in decreasing order of extent. A long-term field trial is planned in the mid-term future, once some further extensions have taken place. The purpose of this field trial will be to provide concrete evidence for our main underlying experimental hypothesis: that human-robot relationships can be significantly enhanced if the human and the robot are gradually creating a pool of shared episodic memories that they can co-refer to ("shared memories"), and if they are both embedded in a social web of other humans and robots they both know and encounter ("shared friends"). Also, notice that for the case of an ongoing project where additions and modifications are still taking place, the evaluations carried out often also function as tuning sessions for parameters or design choices.

A considerable amount of effort was directed towards questions realted to our vision system. A first choice that had to be made was the choice of an appropriate threshold of variance across classifier scores in order to decide that a face should be classified as "unknown", as well as a minimally acceptable winning match score. The underlying assumption here is that an "unknown" face should not create a clear winner among our classifiers, and that even if it does, the corresponding score will be low. The appropriate variance value that was chosen was 1.2 (Fig. 2).

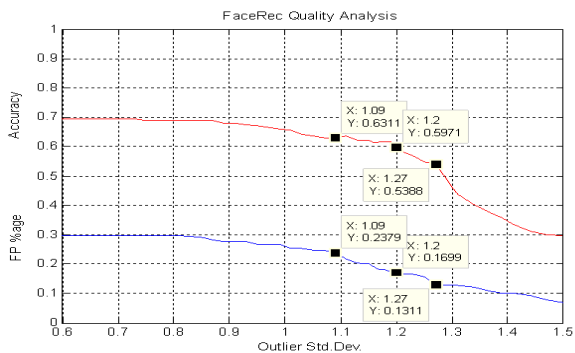

Fig. 2. Effect of varying outlier standard deviation on recognition accuracy and false positive percentage

A second very interesting question is: "over how many frames (window size) should we accumulate evidence before deciding upon the identity of a face?", and also "at what stage and through what accumulation process?". For the latter, after some initial experimentation we decided to accumulate at the level of the continuous scores of the classifiers (before choosing a discrete "winner"), and to do so with a fixed-size-window equal-weight averaging. For the former, we performed empirical tuning, by varying window sizes, and looking at the changes of recognition accuracy. Of course, there is a practical limit to the number of frames; the human does not wait for too long. In practice, a camera-derived training set for five individuals, with a duration of 100 frames was acquired within our lab, and also two testing sets, each of 100 frames, one within the lab (easier), and one outside, where lighting conditions and background differ (more difficult). Then, the window size was varied, and overall accuracy graphs plotted (Fig. 3). From the results, it became clear that after 25 frames or so, corresponding to 5 seconds at 5 frames per second (an acceptable exposure time), there was no more significant increase in accuracy to be expected.

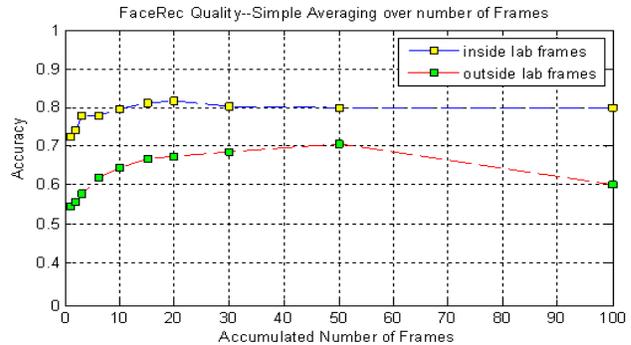

Fig. 3. Recognition Accuracy as a function of evidence accumulation over time (number of frames at 5fps)

A third important question is concerned with the size of the training set. If there is no option for incremental retraining (which is the case in certain recognition methods), and if offline retraining does not take place during idle periods, then one needs to keep retraining time to reasonable lengths. Also, a larger training set is not necessarily a better one; the mid-term variations of human faces (facial hair, movement of light sources) would require more emphasis to be put on the recent shots of the face as compared to older ones. Furthermore, pruning of outliers is another idea we are exploring. Regarding the question of training time versus set size, empirical results are shown in Fig. 4. Thus, 30 or so is the maximal tolerable size for quasi-real-time online retraining (20 seconds), while figures as high as 400 seem to be acceptable for offline (during idle periods), possibly also with a fixed CPU time-slice allocation for backgrounding (15 minutes at 100 percent CPU utilization, one hour at 25 percent - possibly multiplied by the number of people whose classifiers need to be updated).

Now, let us examine the fourth, and most intriguing question. This is more of a system-level question, as it requires interoperation of multiple modules. Having access to both live as well as stored camera pictures, but also to potentially partially or fully tagged Facebook photos, creates many interesting possibilities for using one or the other or a mix for training, and then transferring the knowledge to testing to any of the three species (for automating tagging for example - note that social information can also be utilized in addition towards that purpose, as discussed later in this paper, in the Discussion section). Thus, a crucial question that was asked is: how well do either Facebook photos or camera pictures function as a training set, when they are tested on Facebook photos or camera pictures or across / in a mix? An evaluation was carried out with five persons, and using two 30-picture per person training sets (one from Facebook, and one from the robot vision system detected face regions), and two 30-picture per person disjoint testing sets (again one from each source). The results can be seen in the two across-set accuracy matrices: the difference between that two is that the first one, shown in Fig. 5, uses a camera training set containing photos from five sessions spaced over a month (i.e. has been friend for a while), while the second (Fig. 6)contains a training set containing frames from a single encounter (new friend, have just met once).

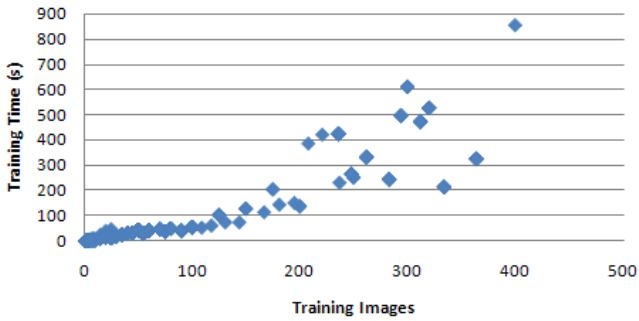

Fig. 4. Time taken for training a classifier

|  | Testing Set: | | |
|---|---|---|---|
| Training Set: | Camera (30) | Facebook (30) | Cam+FB (60) |
| Camera (30) | 98.9 | 46.7 | 72.8 |
| Facebook (30) | 47.8 | 78.9 | 63.3 |
| Cam+FB (30) | 96.7 | 78.9 | 87.7 |
| Cam+FB (60) | 94.4 | 80 | 87.2 |

Fig. 5. Transferability of training from Facebook pictures to camera photos and vice-versa: Recognition Accuracy over different combinations of training and testing sets. Camera training set acquired across one month

|  | Testing Set: | | |
|---|---|---|---|
| Training Set: | Camera (30) | Facebook (30) | Cam+FB (60) |
| Camera (30) | 76.7 | 31.1 | 53.8 |
| Facebook (30) | 48.9 | 78.9 | 63.8 |
| Cam+FB (30) | 75.6 | 77.8 | 76.6 |
| Cam+FB (60) | 73.3 | 74.4 | 73.8 |

Fig. 6. Transferability of training from Facebook pictures to camera photos and vice-versa: Recognition Accuracy over different combinations of training and testing sets. Camera training set acquired from a single session

The results are quite interesting. First, as expected, the temporally spread camera training set (Fig. 5) is much stronger (produces much more accurate recognition) than the single session-derived set (Fig. 6). Second, in the case of the combined camera plus Facebook training sets, one can see that increasing the size of the set from 30 to 60 does not necessarily produce better accuracy - i.e. 30 is adequate (compare row 3 with row 4 in each of the two figures). Third, the cross-combinations (training by camera and testing on Facebook and vice-versa) produce unacceptable results. Camera to Facebook gives 46 percent and 31 percent, while Facebook to camera gives 48 percent and 49 percent - which is noticeably a little better. One could try to explain this asymmetry on the basis of the greater variance across Facebook pictures, which when used as a training set reduces overfitting. On the other hand, Facebook to Facebook performs tolerably (80 percent), and so does camera to camera after a single session. At the top level, we have the camera to camera combination for the case of the spread-over-the-month camera training set, with a high 98 percent. Finally, it is worth noting that although on their own the Facebook-only and camera-only sets do not cross-generalize, if they are used in conjunction (i.e. the third and fourth rows), then they always produce better results than alone in the across-case, and do not significantly deteriorate recognition in the same-species case (98 percent falls to 97, and 76 to 75 etc.).

Finally, notice that at least for the case of real-time camera-shot recognition, errors can be corrected by appropriate verbal feedback by the user; and also, quick experimentation showed that in most cases, in the case of an error, the second-best choice of our system is correct. Thus: "R: Hello! Are you John? H: No R: Oh sorry! I misrecognized you. You are George, right?" is a viable option as it is not so disturbing for your friend to be misrecognized sometimes, as long as he is a new friend (case of Fig. 6 - for a new friend, the robot only has a single session of face training data), and as long as your second guess is correct.

Finally, an initial task-level evaluation was carried out, during which the robot interacted with five people, each one of which for four times. The interactions were videotaped, automated logs were taken, and observations are being analyzed. The duration of the interactions is on the order of 125-145 seconds, during which 8-10 conversational turns are taking place.

## VI. DISCUSSION AND EXTENSIONS

Multiple Extensions are currently underway:
E1) Extensions of the main controller-scripted basic cycle are undergoing testing and further development.
E2) Corresponding language models in order to support the various stages of the robot-driven dialogue are being created. Also, the possibility of supporting some human-initiative or mixed-initiative dialogue turns is considered.
E3) Different ways to utilize the available social information in the form of verbal interactions are being thought out, as well as ways to implement the acquisition of such information through questions.
E4) Increased exploitation of the Facebook messaging and chat channels for verbal interactions is underway, including the possibility of the robot sustaining a conversation with two friends at the same time: one face-to-face physical, and one over Facebook.

E5) The social database and interaction database are undergoing redesign aiming towards ontological (in the philosophical sense) compactness and merging.

E6) Experiments regarding the periodic retraining of classifiers and training set pruning / augmentations are taking place.

E7) A form of basic "active sampling" technique for acquisition of multiple face poses through intentional movement of the pan-tilt of the robot's camera and/or robot body movement is being examined.

E8) The utilization of other online resources apart from Facebook towards driving dialogues and enhancing interactions is being examined.

E9) The possibility of using social-information-driven dialogue as a mode of dialogue, existing alongside other modes (for example, dialogue about the physical situational context, along the lines of [7], and better integration with situation model theory.

E10) The whole system is being moved to a different embodiment: IbnSina, our humanlike humanoid robot (Fig. 7), which supports facial expression, hand gestures, and much more. Also, the system will be possibly extended to a second language.

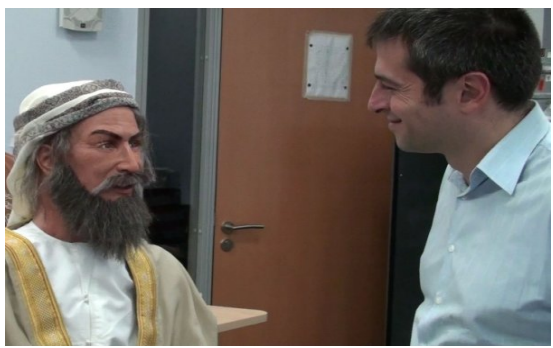

Fig. 7. IbnSina Humanlike Humanoid Robot at the Interactive Robots and Media Lab, UAEU

Furthermore, another extension direction deserves special attention. There exists a possibility for utilizing friendship information, in order to enhance automated tagging in Facebook pictures. The underlying assumption is that friends are more likely to co-occur in photos - thus we can start biasing our recognition hypothesis set towards friends, once we know the identity of a person in a photo. The process goes as follows: suppose we know the identity of a person, either through recognition, or through pre-tagging, and that we are quite confident of it. Then, we acquire his circle of friends through the social database, and we bias our hypothesis space (bigger priors, larger score weight etc.) towards the circle of friends. Then, we recognize the other faces, and choose the one whose identity we are most confident of. Now, we have two circles of friends: the first face's friends (F1), and the second (F2). We also have their intersection: their mutual friends (F1&2). Thus, we can now bias with three levels of strength: small weight for non-friends (not belonging to either F1 or F2), large weight for mutual friends (F1&2), and intermediate weight for friends of F1 or friends of F2 which are not mutual. More implementation details as well as results will be published as this direction unfolds.

Finally, it is worth noticing that the correlation between friendship and co-occurrence in pictures can be utilized in the inverse way too; once we have seen two people co-occuring in multiple photos, it is quite likely that they might be Facebook friends. This idea has been partially utilized for example in the "click-expansion" option of touchgraph [20], and can also provide an indirect way for the robot to have a starting set of hypothesis in order to ask questions to people regarding their friendships.

Now, after having discussed extensions that are underway, we will take a higher-level viewpoint, and discuss where this project fits within a bigger picture. First, this is arguably the first example of utilization and publishing of online information deposited by non-expert humans by an interactive conversational robot, and we foresee a wide array of prospects arising through this stance. Second, since multiple "Facebots" can share social information among themselves, and can "switch embodiments", effectively creating a single identity with multiple distributed embodiments, this creates the prospect for an ultra-social robotic being, which might have a circle of friends much wider than a usual human, and such a being can be beneficial to society in numerous ways. Third, although here we are presenting a mobile robot which can explore physical space and encounter humans, one could easily port a part of the system's functionality to an entity having possibly a virtual body but connected to a physical camera and speech subsystems, effectively remaining stationary in physical space, or anyway just being human-transported. This possibility would also enable a much wider deployment of the system in the near future, which would contribute towards the acquisition of a much bigger training set for refinements and cumulative experience of interactions for analysis.

Last but not least, taking an even higher viewpoint, this is an example of an autonomous system that can have physical or virtual body instantiations, and which can communicate in natural language (and to a certain extent with humanlike manners) with other artificial or biological entities, through physical or electronic channels - and thus we believe it is an important example for theoretical study and speculation towards the new social ecology of the world to come. And exactly there lies our ultimate goal: creating the conditions that would enable harmonious symbiosis of natural and artificial beings across the physical and virtual realms.

## VII. CONCLUSION

Towards sustainable long-term human-robot relationships, a mobile robot with vision, a dialogue system, a social database and a Facebook connection was created, which achieves two important novelties: being the first such robot that is embedded in a social web, and being the first robot that can purposefully exploit and create social information that is available online.

Many side-gains, extensions, as well as a long-term evaluation of our main hypothesis are underway, and we hope that we have brought our ultimate goal of human-robot symbiosis a step closer.

ACKNOWLEDGMENT

We would like to thank Microsoft External Research for supporting this project through its Social Robotics CFP 2007